\documentclass[10pt, a4paper]{article}

\usepackage{lrec-coling2024} 
\usepackage{times}
\usepackage{latexsym}
\usepackage{graphicx}
\usepackage{amsmath}
\usepackage{booktabs}
\usepackage{tabularx}
\usepackage{multirow}
\usepackage{listings}
\usepackage[raggedrightboxes]{ragged2e}
\usepackage{microtype}
\usepackage{inconsolata}

\title{\textsc{ReflectSumm:} A Benchmark for Course Reflection Summarization}

\name{Yang Zhong\textsuperscript{$\dagger$}*, \thanks{\text{  }\text{  }\text{  }* These authors contributed equally to this work.}Mohamed Elaraby\textsuperscript{$\dagger$}*, Diane Litman\textsuperscript{$\dagger$}, \\ {\bf \large Ahmed Ashraf Butt\textsuperscript{$\clubsuit$}, Muhsin Menekse\textsuperscript{$\diamond$}}
} 

\address{\textsuperscript{$\dagger$} Department of Computer Science, School of Computing and Information\\
         University of Pittsburgh, Pittsburgh, USA \\
         \textsuperscript{$\clubsuit$}
School of Computer Science, Carnegie Mellon University, Pittsburgh, USA\\
         \textsuperscript{$\diamond$} School of Engineering Education, Purdue University, West Lafayette, USA\\
          \{yaz118, mse30, dlitman\}@pitt.edu \\
        ahmedasb@cs.cmu.edu\\ menekse@purdue.edu\\}

\abstract{
This paper introduces \textsc{ReflectSumm}, a novel summarization 
dataset specifically designed 
for summarizing students' reflective writing. The goal of \textsc{ReflectSumm} is to facilitate developing and evaluating novel summarization techniques tailored to real-world scenarios with little training data,
with potential implications in the opinion summarization domain in general and the educational domain in particular. The dataset encompasses a diverse range of summarization tasks and includes comprehensive metadata, enabling the exploration of various research questions and supporting different applications. To showcase its utility, we conducted extensive evaluations using multiple state-of-the-art baselines. The results 
provide benchmarks for facilitating further research in this area.  \newline  }

\begin{document}

\maketitleabstract

\section{Introduction}

Advances in Pretrained Language Models 
\cite{raffel2020exploring, lewis-etal-2020-bart, zhang2020pegasus} and Large Language Models \cite{brown2020language, chowdhery2022palm, scao2022bloom, touvron2023llama} have propelled neural summarization to new heights. 
Existing research has primarily focused on standard summarization benchmarks within domains like news \cite{hermann2015teaching, narayan2018don}, dialogue \cite{gliwa2019samsum}, scientific articles \cite{cohan2018discourse}, and opinions \cite{angelidis2018summarizing,  chu2019meansum, bravzinskas2020few}. However, there is 
also a need for 
benchmarks that better represent real-life applications of summarization. These benchmarks should explore areas that have received limited attention 
and that present new and challenging use cases. By incorporating these underexplored domains into the evaluation process, we can effectively assess the performance of summarization models in scenarios where summarization can make a meaningful social impact.
This paper addresses this need by introducing \textsc{Reflectsumm}, a novel dataset 
focusing on the summarization of 17,512 student reflections on 782 university lectures from 24 large STEM classes. Table \ref{example} shows example  
reflections in response to a prompt regarding the interesting facets of a lecture.
\begin{table}[h]
\footnotesize
\begin{tabular}{p{0.44\textwidth}}
\toprule
\textbf{Reflection Prompt}\\ Describe what you found most interesting in today’s class   \\ \midrule
\textbf{Student Reflections}\\ 
{\scriptsize $\bullet$ Nothing in particular today -> 1.0} \\ 
{\scriptsize $\bullet$ Despite the confusion, I did find setting up these problems to be very interesting and rewarding. -> 3.0} \\
{\scriptsize $\bullet$ Equipotentials -> 2.0} \\
{\scriptsize $\bullet$ i thought the breakout room questions were interesting because i learned how to do questions -> 4.0} \\
{\scriptsize $\bullet$ I found the last problem in class the most interesting because it was proven we can derive almost anything. -> 4.0 }\\
{\scriptsize $\bullet$ The most interesting thing was that finding electric potential doesn't require a path, but only the magnitude of the charge and it's distance from the point of interest. -> 4.0} \\
{\scriptsize $\bullet$ I really enjoy line integrals and I can tell that we're moving towards using them to calculate potential. -> 4.0} \\
{\scriptsize $\bullet$ Collection of point charges (pairing them) -> 2.0}\\
{\scriptsize $\bullet$ How we can calculate something so complicated as electrons passing through an area is very cool. -> 3.0} \\
{\scriptsize $\bullet$ I found equipotentials to be the most interesting thing, especially drawing a equipotentials for a dipole! -> 4.0} \\
{\scriptsize $\bullet$ I thought it was interesting that Vnet is equal to all Vs added together -> 4.0} \\
{\scriptsize $\bullet$ I found how conductors act to be interesting. -> 3.0 }\\

\textit{ \textbf{...} the rest is omitted to save space} \\
\midrule

\textbf{Abstractive Summary}\\ The students today found calculations and relationships to other concepts that they have learned in this and other classes interesting. They also found potential energy and equipotentials very interesting, as well as some integration concepts. \\ \midrule
\textbf{Extractive Summary}\\
$\bullet$ I found equipotentials to be the most interesting thing, especially ...\\
$\bullet$ The most interesting thing was that finding electric potential doesn’t require a path, but only the magnitude of the charge and it ...\\
\textit{... three more extractive reflections omitted to save space} \\
\midrule
\textbf{Phrase Summary}\\ $\bullet$ equipotentials\\ $\bullet$ calculations\\  $\bullet$ relations to old concepts \\$\bullet$ potential\\$\bullet$
integration \\
\bottomrule

\end{tabular}
\caption{An example from the 
\textsc{ReflectSumm} dataset
 showing reflections annotated with specificity score (displayed after the special token ``->'') and three different types of reference summaries.\label{example}}
\end{table}
As suggested by \citet{Baird1991reflection}, reflections are useful  
for both students and teachers, 
enhancing their knowledge, self-awareness, and classroom practice. For example, providing  reflection summaries can assist instructors in identifying key areas where students exhibit misconceptions, thereby enabling them to strategize appropriate follow-up actions for upcoming lectures \cite{ fan-etal-2017-reflectionprompt}. Compared to using human-crafted summaries, {\it automatic summarization can help scale the use of 
reflections in educational practice}. 


It is important to recognize that {\it student reflections and their summaries differ from standard benchmark 
corpora} in the related area of opinion summarization,\footnote{Opinions
are similarly obtained from multiple humans and order doesn’t matter.} which has traditionally focused on 
product and service reviews. Table \ref{example} illustrates the {\it variability observed in the length and structure of 
 reflections}.  While some students opt for concise expressions using words or phrases, others delve deeper into the topic by composing complete sentences to highlight interesting lecture aspects.  {\it Reflection summaries are also more abstractive} than standard opinion summaries (see Table \ref{data_stats}, to be discussed below). 

 Furthermore, {\it {\sc ReflectSumm} provides richer types of information compared to 
existing corpora} for summarizing 
both student reflections  \cite{ luo-etal-2016-automatic,  fan-etal-2017-reflectionprompt,Magooda2020AbstractiveSF} as well as opinions \cite{angelidis2018summarizing, bravzinskas2020few, angelidis-etal-2021-extractive, Yang2022OASumLO}. While prior corpora emphasized either abstractive or extractive summarization, our dataset provides {\it three types of reference summaries} for each set of reflections: extractive, abstractive, and phrase-level extractive summaries. 
Additionally, we augment the dataset with valuable metadata, such as reflection {\it specificity scores},\footnote{Each reflection in Table \ref{example} is assigned a score ranging from 1 to 4 (explained in Section \ref{data_desc}).} which can be used to improve summarization performance 
(see Section \ref{sec:result_and_analysis}). Furthermore, we provide student {\it demographic information}, enabling the exploration of fairness and equity issues. 

Our contributions can be summarized as follows:
$(1)$ We publicly release 
\textsc{ReflectSumm}, 
which contains 17,512 reflections on 782 lectures from 24 university courses, along with reference summaries and 
metadata,  allowing for 
exploration and advancement in 
summarization. 
$(2)$ We conduct a detailed analysis using both pretrained language models and large language models 
to benchmark the \textsc{ReflectSumm} dataset across abstractive, extractive, and phrase summarization tasks.
$(3)$ We investigate research directions leveraging the provided metadata by exploring the concept of \textit{specificity-aware summarization}. 
The specificity metadata provides a way to integrate the study of specificity \cite{jessy-2015-specificity, Gao_Zhong_Preoţiuc-Pietro_Li_2019} into summarization. Additionally, we showcase that our demographic information can assist further research in studying the fairness and bias problem in the context of summarization research (Sections \ref{data_desc} and \ref{sec:broader_impact}).
$(4)$ We make our dataset, models, and model outputs publicly available at \url{https://github.com/EngSalem/ReflectSUMM}, enabling researchers to build upon our work. 

\section{Related Work}
Prior student reflection datasets \cite{luo-litman-2015-summarizing, luo-etal-2016-automatic,Magooda2020AbstractiveSF} 
were constrained in their size, course diversity,  and summarization task coverage (see Table \ref{tab: reflect_datasets}, to be discussed below).   
 Specifically, prior datasets not only summarized fewer lectures, 
 but also covered fewer academic subjects and/or courses per subject, limiting investigations of how models generalize.  
 In addition, only one of our three summarization tasks (extractive, abstractive, and phrase-based) is covered per prior work. 
Well-known review opinion summarization benchmarks are similarly constrained in their summarization task coverage, with OpoSum \cite{angelidis2018summarizing}  focused on extractive summarization and FewSumm \cite{bravzinskas2020few} instead focused on abstractive summarization (Table \ref{data_stats}). 
\textit{\textsc{ReflectSumm} 
provides reference summaries in three formats  (abstractive, extractive,  phrase-based), 
 new types of metadata (reflection-level specificity annotations, 
 student demographic information), and 
enables various evaluation scenarios (including but not limited to cross-course, within-course, course-agnostic, cross-subject, etc.).} 
Most prior NLP work on student reflections has focused on quality (e.g. specificity) prediction 
\cite{Kovanovic2018UnderstandingSelfReflection, Ullmann2019AutomatedAnalysis, Carpenter2020MiddleSchoolReflection}. With respect to summarization, \citet{luo-litman-2015-summarizing} suggested extracting noun phrases to 
 compress the reflections for supporting mobile applications.
\citet{Magooda2020AbstractiveSF} 
utilized neural models with a focus on generating abstractive summaries in a low-resource
context 
\cite{magooda-etal-2021-exploring-multitask, magooda2021mitigating}. 
\textit{Our work follows the neural paradigm, further developing prior pretrained baselines and
exploring the use of Large Language Models (LLMs).}


Our dataset can be considered as a special case of low-resource multi-document opinion summarization, where the opinions here refer to student reflections 
rather than service and product reviews \cite{angelidis2018summarizing, bravzinskas2020few}. Most prior opinion work with limited data focused on synthesizing training data for intermediate finetuning \cite{bravzinskas2020few},  parameter efficient techniques \cite{bravzinskas2022efficient}, or second stage reranking \cite{oved2021pass}. Previous low-resource work that targeted summarizing both student reflections and product reviews leveraged multitask learning with pretrained language models \cite{magooda-etal-2021-exploring-multitask}, domain transfer from pretrained models \cite{Magooda2020AbstractiveSF}, and curriculum learning \cite{magooda2021mitigating}. 
Recently, Large Language Models 
\cite{brown2020language, scao2022bloom, sanhmultitask, touvron2023llama} 
have been explored for both news \cite{goyal2022news,zhang2023benchmarking} and opinion \cite{bhaskar2022zero} summarization in  zero-shot settings. \textit{We provide several baseline results with both pretrained language models and LLMs to benchmark their utility in the zero-shot and one-shot summarization of reflective writing.}
\section{\textsc{ReflectSumm}}
\label{sec:reflectsumm}
\subsection{Dataset Collection and Annotation}
\label{subsec:data_collection}

The student reflections in \textsc{ReflectSumm} were collected after each lecture in 24 
courses from two American universities. The data were obtained across four semesters, from Fall 2020 to Spring 2022. Students used the CourseMirror Application \cite{Fan2015COURSEMIRROR}\footnote{The application is downloadable from Apple Store \url{https://apps.apple.com/us/app/coursemirror-v2/id1506495976} and Google Store \url{https://play.google.com/store/apps/details?id=education.pittsburgh.cs.mips.cm_v2&hl=en_US&gl=US&pli=1}} to respond to two 
prompts: 
\textit{(1) Describe what you found most interesting in today's class
} and \textit{(2) Describe what was confusing or needed more details in today's class}.    These reflection prompts are based on learning sciences research, starting with \citet{menekse2011reflection}, where students wrote reflections on paper and a TA manually summarized them. These 
prompts are polarity-specific (confusing versus interesting lecture aspects). Prior evaluations of early versions of the CourseMIRROR app used to collect our data (where the app only used phrase summarization at the time) found that reading the phrase summaries was viewed positively by both instructors and students \cite{Fan2015COURSEMIRROR}. More recently, \citet{menekse-2020-reflection} found that generating reflections and reading class phrase summaries improved student exam scores.

\begin{figure}[!ht]
\begin{center}
\includegraphics[scale=0.6]{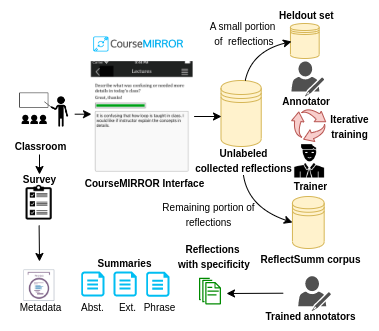}  \caption{\centering  \textsc{ReflectSumm} corpus creation.} \label{fig:data_annotations}
\end{center}
\end{figure}

For the reflection quality annotation, following the guidelines in \citet{Luo2016DeterminingTQ},  annotators assigned a score from 1-4, where 4 means the reflection text has the highest specificity and 1 means the least specificity.

 Turning to summarization, the 
phrase summary task for the annotator is to provide five phrases that can best summarize the students' reflections, together with how many students semantically mentioned each phrase. Those phrases can be either extracted from the reflections or manually constructed by the annotator. 
The annotators are further instructed to write an abstractive summary to summarize the major points of the full reflections. Lastly, the annotators select five reflections as the extractive summaries. Full annotation guidelines are in Appendix \ref{sec:appendix_annotation_guideline}.


Eleven college students with backgrounds in the appropriate subject domains were recruited to work on 
 reflection scoring and summarization.  
 Students were first trained on three batches of extra-held sets to understand and grasp the tasks before being assigned real jobs. 
The average pairwise inter-annotator agreement (IAA) across four students with double-annotations
is 0.668 for the reflection score by Quadratic Weighted  Kappa, suggesting substantial agreements. For summarization, 
we instead measure the averaged inter-annotator ROUGE scores (R-1/R-2/R-L) \cite{lin2004rouge}, which are  48.31/27.57/43.52 and 30.16/6.77/27.91 for the extractive and abstractive summarization tasks, respectively. These scores are slightly lower than  
those reported in \citet{pittir42259}, which used the same guidelines but on different data. Figure \ref{fig:data_annotations} provides a summary of the annotation process involved in constructing \textsc{ReflectSumm}.\footnote{We justify the reason to select undergraduate students as annotators in Appendix \ref{sec:appendix_human_selection}.}




\begin{table*}[ht!]
\small
\centering
\setlength{\tabcolsep}{4pt}
\begin{tabular}{lcccccc}
\toprule
\textbf{Dataset} & \textbf{\# Pairs} & \textbf{\# Words/input} & \textbf{\# Docs/input} & \textbf{Abstractiveness} & \textbf{Tasks} \\
& 
& (min/avg/max) & (min/avg/max) &  (1/2/3-grams) & \\
\midrule
\textbf{OPOSUM} & 600 & 468/485/499 & 10/10/10 & - & Ext. \\
\textbf{\textsc{FewSumm} Amazon} & 180 & 342/397/438 & 8/8/8 & 25.02/78.29/97.65 & Abst. \\
\textbf{\textsc{FewSumm} Yelp} & 300 & 362/399/433 & 8/8/8 & 26.04/80.71/98.76 & Abst. \\
\midrule
\textbf{\textsc{ReflectSumm}} & 782 & 10/344/2229 & 4/22/79 & 36.97/83.11/98.12 & Abst./Ext./Phrase \\
\bottomrule
\end{tabular}

\caption{Descriptive statistics comparing prior datasets (top) to \textsc{ReflectSumm}. \textbf{\# Pairs} denotes the number of reflection/review document and summary pairs, while \textbf{\# Words/Input (concatenated reflections/reviews)} represents the total word count in the input. \textbf{\# of Docs (reflections/reviews)/input} indicates the number of documents per input . \textbf{Abstractiveness} measures abstractive summaries' novelty in terms of n-grams. (-) signifies that the information is not applicable for this dataset. \textbf{Tasks} include Abstractive (Abst.), Extractive (Ext.), and Phrase (Phr.) summarization. \label{data_stats}}
\end{table*}

 \begin{table*}[ht!]
\small
\centering
\setlength{\tabcolsep}{5pt}
\begin{tabularx}{\textwidth}{lp{1.3cm}p{4cm}p{3cm}p{1cm}}
\toprule
\textbf{Dataset} &  {\textbf{\# Pairs 
}} & \textbf{Course Coverage} & \textbf{Tasks} & \textbf{Metadata} \\
\midrule
\citet{luo-litman-2015-summarizing} &	36 & 	1-ENGR	& Phrase & No\\
\midrule
\citet{luo-etal-2016-automatic} &	70 	 & 2-Statistics	& Phrase & No \\
\midrule
\citet{Magooda2020AbstractiveSF}	& 188 &	4-ENGR/Statistics/CS	& Abst. & No \\
\midrule
	\textbf{\textsc{ReflectSumm} (ours)}  & 782	& 24-ENGR/PHY/CS/CMPINF	& Phrase/Abst./Ext. & Yes \\


\bottomrule
\end{tabularx}

 \caption{Comparison of \textbf{\textsc{ReflectSumm}} 
 with previous reflection summarization datasets. \textbf{\#Pairs} represents the count of unique lectures featuring the "interesting/confusing" prompt in the summarization input.  \textbf{Course Coverage} describes the sources of reflections, presented in a count-course subjects format, \textbf{Tasks} delineates the 
 types of reference summaries, and \textbf{Metadata} indicates the inclusion of additional information in the corpus (e.g., student demographics,  reflection specificity). 
 \label{tab: reflect_datasets}}

\end{table*}

\subsection{Dataset Description and Details}\label{data_desc}
\textsc{ReflectSumm}  reflections were collected after 782 lectures from 24 courses spanning four different subjects: Engineering (ENGR), Physics (PHY), Computer Science (CS), and Computing Information (CMPINF).\footnote{ENGR courses come from a midwest US university, while the rest come from a northeast university.}
 The majority of both lectures $(56.9\%)$ and courses $(54.2\%)$ are from ENGR. The remaining data is fairly evenly distributed between CS ($20.3\%$ of lectures and $20.8\%$ of courses) and PHY ($18.2\%$ of lectures and $16.7\%$ of courses), with only a small percentage of CMPINF lectures and courses ($4.6\%$ and $8.3\%$, respectively). 

A total of $17,512$ \textsc{ReflectSumm}  reflections spanning the $782$ lectures have been annotated for their specificity as noted above. The majority of the reflections $(52.5\%)$ were rated as having a specificity score of 3, indicating moderate specificity.  $22.6\%$ of the reflections received the highest  
rating of $4$, 
while 
$14.2\%$ received a score of $2$
and only $10.7\%$  
received a score of $1$. 
The average specificity score of sentences selected for extractive summaries is higher (3.08) compared to the discarded sentences (2.85), 
suggesting that considering specificity 
as additional information to guide summarization 
is worth exploring.
 
Before the data collection phase, students were asked to participate in a pre-survey that collects their demographic information. Male students make up the majority $(55.71\%)$, while the survey did not record $3.45\%$ of students' gender information. 
We also observe a diverse distribution across multiple racial groups, and the majority (58.89\%) are White 
followed by Asian 
(20.93\%). 

Table \ref{data_stats} compares \textsc{ReflectSumm} with several established multi-document opinion summarization datasets. 
\textbf{\textsc{ReflectSumm}} boasts 
782 input-summary pairs, which represent the count of unique lectures featuring the "interesting/confusing" prompt in the summarization input. This dataset surpasses 
OPOSUM \cite{angelidis2018summarizing}
and FewSumm \cite{bravzinskas2020few}  in terms of dataset size (column 2).
While 
OPOSUM and FewSumm limit the number of documents per input, \textsc{ReflectSumm} has more variability in 
the number of words and documents per input (columns 3 and 4). 
Column 5 shows that our abstractive summarization task is more 
abstractive, as measured by the percentage of novel n-grams \cite{see-etal-2017-get}.\footnote{novelty =  \texttt{\#} new n-grams in summary/  \texttt{\#} total n-grams in summary} 
 Lastly, \textsc{ReflectSumm} 
 encompasses a distinctive blend of 
 summarization tasks (column 6). 
%
 Table \ref{tab: reflect_datasets} compares \textsc{ReflectSumm} with prior reflection-focused corpora. Columns 2-5 show that our dataset surpasses in the number of unique lectures featured with focused prompts, 
the breadth of courses covered,
the diversity of reference summaries, and the availability of metadata. 

\section{Tasks and Benchmark Models}


\subsection{Extractive Summarization}
Corresponding to the human extractive summary task, the goal of our models is to pinpoint the five most salient reflections (documents) from a collection of reflections within the same lecture.  
We evaluate several baseline models to benchmark extractive performance:  the traditional unsupervised method, \textbf{LexRank} \cite{erkan2004lexrank}; \textbf{BERTSUM-EXT} \cite{liu-lapata-2019-text}, a pretrained language model crafted for extractive summarization; \textbf{MatchSum} \cite{zhong-etal-2020-extractive}, a state-of-the-art  summarization system which employs a re-ranker, and follows a two-stage paradigm to extract summaries; and \textbf{ChatGPT (GPT-3.5 turbo)}, a large language model capable of generating high-quality summaries. For ChatGPT, we devised two variants of prompts for experimentation in a zero-shot setting: 
\textbf{($1$) GPT-reflect:} This variant prompts the model to select complete reflections, which mirrors how students typically write their reflections.
\textbf{($2$) GPT-reflect + specificity:} This variant integrates specificity scores with the original reflection and prompts the model to consider these scores while making selection choices. 
We also created a one-shot setting for the best-performing zero-shot model by randomly selecting a 
training split example and instructing the model to follow the example.

\subsection{Extractive Phrase Summarization}
The extractive phrase summarization task seeks to generate summaries constituted by five phrases, each supplemented with an accompanying number that indicates how many reflections support each phrase. This numerical task is inspired by the desire for a more comprehensible display of reflection distributions to instructors  \cite{Fan2015COURSEMIRROR} while phrase extraction facilitates easier access on mobile devices. To benchmark phrase summarization, we have employed the unsupervised model proposed by \citet{luo-litman-2015-summarizing}, aggregating noun phrases into five clusters and identifying the clusters' centroids as the phrase summary. We named it \textbf{PhraseSum}.
In addition, we utilized OpenAI's \textbf{ChatGPT (GPT-3.5-turbo)} as our LLM to perform zero-shot phrase extraction. We experimented with \textbf{GPT-Human}, using the prompt provided to human annotators.  However, as the original prompt doesn't specify the types of phrases to extract, we introduced two more baselines for additional experiments: (1) \textbf{GPT-noun phrase} aimed to extract just noun phrases for each lecture, and (2) \textbf{GPT-Human + noun} which adds  ``noun phrase'' to the original human prompt. 
We also added a one-shot setting for the best-performing model, similar to the extractive summarization task.
We include all prompts in Appendix \ref{sec:appendix_all_prompt}.

\subsection{Abstractive Summarization}


Human annotators were given the task of summarizing students' reflections concisely and coherently within $~40$ words. To benchmark this task, various models were employed, including fine-tuning pretrained language models, namely \textbf{BART-Large} \cite{lewis-etal-2020-bart} and a modified version called \textbf{BART-Large+specificity}. 
 The latter incorporates markers on a 4-point scale indicating reflection specificity scores, following the same approach used in literature for scientific articles \cite{deyoung2021ms2}, dialogue \cite{khalifa2021bag}, and legal documents \cite{elaraby2022arglegalsumm}.\footnote{See Appendix \ref{sec:appendix_abst_summ} for an example of using markers to include specificity scores.}


As with the extractive models, we developed ChatGPT models too. 
In the zero-shot setting, we explored two prompting settings: \textbf{($1$) GPT-Human:} using a  version similar to  the prompt given to human annotators
and \textbf{($2$) GPT-Human + specificity:} incorporating specificity scores of each reflection in the prompt. 
We also added a one-shot model 
as in the extractive summarization tasks.

\section{Experimental Setup}

All models are evaluated using cross-validation. Lectures are first grouped and shuffled by the subjects.  Each subject is then divided into five folds by 
shuffling the lectures within that subject. We combine those four folds from each subject as the final training fold set and make the remaining test fold. We randomly select $10\%$ of the data within each training fold set for validation and model selection.

We mainly evaluate our models using two standard metrics, ROUGE \cite{lin2004rouge} and BERTScore \cite{zhang2019bertscore}. In addition, we report lecture-level reflection exact match F1 (EM F1) and partial match F1 (P F1) scores for the extractive summarization task.\footnote{See Appendix \ref{appendix:partial_f1_illustration}
for an illustrative example of how these scores are computed.} 
 For the exact match F1, we compare the predicted and human-reference reflections on a per-lecture basis. Partial match F1 assesses the correctness of selecting partial components from a complete reflection, allowing for more flexibility in the evaluation.
 While standard extractive summarization tasks typically use human-written abstractive summaries as references, we utilize the annotated extractive reference summaries to evaluate the model outputs. 


 Abstractive models often suffer from hallucinations, generating information not present in the source text \cite{ji2023survey}. To assess the factuality of our generated summaries, we utilize the pre-existing 
 entailment metric called $\textsc{SummaC}$ \cite{laban2022summac}.
This metric assesses the overall entailment score between the generated summary and the input document, considering different levels of granularity. The score can be computed by considering the entire document 
or computing the aggregated score from pair-wise sentence-level entailments. 
\textsc{SummaC} introduces two versions: \textsc{$SummaC_{zs}$}, which averages pairwise entailment scores. A score (ranging from $0$ to $1$) indicates a stronger alignment between the document and the summary, while a negative score (ranging from $0$ to $-1$) suggests counterfactually-generated text; and \textsc{$SummaC_{conv}$}, where entailment scores are aggregated by a convolution layer to avoid mean sensitivity to extreme entailment values. The convolution layer aggregates values into 5 bins: $[0, 0.2)$, $[0.2, 0.4)$, $[0.4, 0.6)$, $[0.6, 0.8)$, and $[0.8, 1)$. A higher bin indicates a stronger factual consistency of the summary. We mainly relied on \textsc{$SummaC_{conv}$} in our analysis, as recommended by the paper.\footnote{We opted against using factuality metrics based on question-answering (QA) approaches like QAFactEval \cite{fabbri2022qafacteval}. This decision was due to limitations in entity extraction, which struggled to recognize educational concepts and noisy noun phrase generation caused by the varied structure of input reflections, as shown in the example included in Appendix \ref{app:factuality_evaluation}.} 


\section{Implementation Details}
For the BERTSUM-EXT \textbf{ extractive summarization} model, we leveraged a bertext\_cnndm\_transformer checkpoint that was trained on the CNN/DM news dataset using the original codebase\footnote{\url{https://github.com/nlpyang/PreSumm}} to select 5 reflections. We additionally fine-tuned BERT-EXT models on our dataset and the \textsc{FewSumm Amazon} dataset to examine the benefits of our data for summarization tasks. We further experimented with BERT-EXT + specificity, where the specificity scores are incorporated into the input. For MatchSum, we used the checkpoint equipped with a RoBERTa-based re-ranker. We formed the candidate sets by employing the off-the-shelf BERT-EXT model to prune the original documents into 8 reflections and constructed the combinations of 5
sentences subject to the pruned document. For LexRank, we used the lexrank package,\footnote{\url{https://github.com/crabcamp/lexrank}} treating the concatenation of all reflections from the train split of each fold as documents to initialize the model and setting the summary size at 5 with threshold ratio of 0.1. For ChatGPT, 
we utilized the OpenAI API.\footnote{\url{https://platform.openai.com/docs/api-reference}}
We set the maximum tokens to $1024$ and the temperature to $0.5$.

We replicated the \textbf{extractive phrase summarization} model from 
\citet{luo-litman-2015-summarizing}, which utilizes KMedoid clustering of 
noun phrases extracted from 
reflections and encoded using the BERT-base model (details in Appendix \ref{sec:appendix_phrasesum_detail}).
For ChatGPT, 
we set the maximum tokens to $1024$ and the temperature to $0.5$.

We fine-tune the BART-Large \textbf{abstractive summarization} model for 10 epochs on each fold, employing an early stopping technique with the patience of 3 epochs. We utilize 
the HuggingFace implementation \cite{wolf2019huggingface}.
To identify the optimal model, we evaluate its R-2 score on the validation set. For the BART + reflection specificity tokens, we use human-annotated specificity scores during training and predicted specificity obtained by a finetuned-DistillBERT model.
For LLMs, we use openAI's API 
and set the maximum tokens to $100$ and the temperature to $0.7$.   

\section{Results and Analysis}\label{sec:result_and_analysis}

\begin{table*}[]
\small
\centering
\begin{tabular}{l|lll|l|l|c}
 \toprule
\textbf{Model}  &  \textbf{R-1}  &  \textbf{R-2}  &  \textbf{R-L} & \textbf{BS} & \textbf{EM F1} & \textbf{P F1} \\
\midrule
LexRank & 56.96	& 40.45 &	55.10	& {89.95}	& 31.33	& {37.16} \\
\midrule
BERTSUM-EXT (cnndm) &55.21 &	38.29	& 53.39	 & 89.91 & {34.08}	& 37.74\\

BERTSUM-EXT (ft. \textsc{FewSumm Amazon}) & 55.51 & 38.41 &	53.59& 89.88 & 33.85 & 37.81 \\
BERTSUM-EXT (ft. \textsc{ReflectSumm}) & 56.15	& 39.09 &	54.25 &	89.94& 33.14 & 37.49 \\
BERTSUM-EXT (ft. \textsc{ReflectSumm}) + Specificity &55.94	& 39.50 &	54.16 &	89.31 & {33.45}	& {37.84}\\
\midrule
MatchSum & 58.79* &	42.59*	& 56.70* & 	\textbf{90.57} &  \textbf{36.26}* & \textbf{38.94}\\
\midrule
GPT-reflect & \textbf{60.16}*  &	\textbf{43.93}* &	\textbf{58.26}* & 89.98   & 21.41	& {37.68} \\
  GPT-reflect + specificity & {58.76}*  &	{42.49}* &	{56.85}* &  90.29 & 21.28	& {36.45} \\ \midrule
GPT-reflect - one-shot  & 58.65 &	41.04		&56.46	& 89.58 &	20.18	&33.07\\
\bottomrule
\end{tabular}
\caption{Extractive summarization model performance reported on ROUGE (R-1, R-2, R-L), BERTScore (BS), Exact Match F1 (EM F1) and Partial F1 (P F1). The best column results 
are \textbf{bolded}, while * means
statistically different from the
baseline LexRank (p-value < 0.05) using a paired t-test.}\label{tab:extractive_sent_table}
\end{table*}

\begin{table}[]
\small
\centering
\setlength{\tabcolsep}{3pt}
\begin{tabular}{l|ccc|c}
 \toprule
\textbf{Model}  &  \textbf{R-1}  &  \textbf{R-2}  &  \textbf{R-L} & \textbf{BS} \\
\midrule
PhraseSum & 24.87 &	7.98 & 24.31 & 83.9\\
\midrule
 GPT-Human & 34.25* &	11.25*	& 33.27*	& 84.7*  \\
 GPT-noun phrase & {39.28}*	& {14.55}*	& {38.26}*	& {87.1}*\\
 GPT-Human + noun & 38.86* &	13.56*	& 38.02*	&84.6* \\ \midrule
 GPT-noun - one-shot & \textbf{40.43}* & \textbf{15.48}* & \textbf{39.51}* & \textbf{87.7}*\\
\bottomrule
\end{tabular}
\caption{Extractive phrase summarization model performance. The best result of each column is \textbf{bold}. * means
statistically different from the 
baseline PhraseSum (p-value < 0.05) using a paired t-test.}\label{tab:extractive_phrase_summary}
\end{table}

\subsection{Extractive Summarization} Table \ref{tab:extractive_sent_table} shows that the baseline BERTSUM-EXT is not as satisfactory as the traditional LexRank baseline. We observe that fine-tuning the model on our specific dataset brings appreciable performance gains, as evidenced by the comparison between row 2 and row 4.
Furthermore, fine-tuning on a similar opinion summarization dataset also enhances performance, though not to the same extent as using our in-domain data (row 2 vs. row 3). Including specificity scores in the dataset brings improvements in R-2 and helps with the matching F1 when compared to the fine-tuned baseline without specificity information (row 4 vs. row 5). The state-of-the-art model MatchSum obtained the second-best performances regarding ROUGE scores and the highest BERTScore and matching performances.\footnote{Example outputs for all models are in Appendix \ref{sec:appendix_examples}.} 
Meanwhile, ChatGPT models obtain the best or on-par performance
concerning ROUGE and BERTScores. However, it should be noted that the ChatGPT-based models struggle to fully extract the reflections, as evidenced by the lower Exact Match F1 scores. We posit the improvements from Exact to Partial F1 are attributed to the incapability of ChatGPT models to comprehend the prompt fully, thus cutting the original reflections into sentences and making partial selections.
Overall, the Partial F1 score suggests that there is still ample room to improve the extractive summarization models to match human performance. Based on the best zero-shot setting (\textit{GPT-reflect}), we also explored the one-shot setting and found no gains.

To validate the hypothesis that ChatGPT-based models may not be able to perform the extractive task faithfully, we analyzed the proportion of model output sentences that are fully extracted from the original reflections. In detail, we measure the ratio of the system-extracted reflections that come from the original reflections instead of being generated creatively by the ChatGPT model. Compared to the BERTSUM-EXT model with near-perfect extractiveness (99.4\%),\footnote{The score does not reach 100\% since some annotators selected sentences instead of full reflections.} 
ChatGPT models obtained 92.53, and 94.68\% extractiveness scores for \textit{GPT-reflect} and \textit{GPT-reflect + specificity}, respectively, showing that GPT models do sometimes generate non-extractive sentence/reflections. In contrast, MatchSum follows a two-stage paradigm to extract summaries. The output reflections are guaranteed to be selected from the original reflections, securing higher EM F1 and P F1.

\subsection{Extractive Phrase Summarization}\label{sec:extractive_phrase}
Table \ref{tab:extractive_phrase_summary} shows the results.\footnote{We apply regular expressions to clean up the GPT model outputs. Details are in Appendix \ref{appendix:regular_expression}.} 
The PhraseSum model obtains the worst performance. We observe a difference when adjusting the prompt for the ChatGPT models. 
Comparing  \textit{GPT-Human} and \textit{GPT-Human + noun} shows that just replacing  ``phrases'' with ``noun phrases'' brings about 4.5 points improvements on R-1 and R-L, 2.3 points on R-2. Meanwhile, \textit{GPT-noun phrase}  obtains improvements of 5.0, 3.3, and 4.9 ROUGE (1,2, L) scores compared to \textit{GPT-Human}.  We posit that the relatively lower performance of {\it GPT-Human} compared to \textit{GPT-noun phrase} is that, in \textit{GPT-Human}, the task is inherently a multi-task project by adding the prompt of \textit{together with how many students semantically mentioned each phrase in parenthesis}.  Finally, based on the best zero-shot setting, we explored a one-shot setting for \textit{GPT-noun phrase}, where a random sample from the data was used. Unlike in Table~\ref{tab:extractive_sent_table}, one-shot outperforms all zero-shot settings, perhaps because phrase samples are more consistent than the full reflections.
\begin{table}[]
\small
\begin{tabular}{l|ccl|c}
 \toprule
\textbf{Model}  &  \textbf{R-1}  &  \textbf{R-2}  &  \textbf{R-L} & \textbf{BS} \\
\midrule
BART-Large & 47.09 & 24.17 & 43.76 &90.49 \\ 
+ specificity & \textbf{47.70} & \textbf{24.85} & \textbf{44.41*} & \textbf{90.57} \\ \midrule

GPT-Human & 35.83 &	9.40 &	31.85 & 88.23 \\

+ specificity  & 36.73 & 9.13 & 31.64 & 88.27 \\ \midrule

GPT-one-shot & 36.86 & 9.46 & 31.96 & 88.26 \\
\bottomrule
\end{tabular}
\caption{Abstractive summarization performance. Best column result \textbf{bolded}; * is statistically different from the baseline BART-Large.} \label{tab:abst_summ}
\end{table}

\begin{table}[]
\small
\setlength{\tabcolsep}{5pt}
 \begin{tabular}{l|c|c}

 \toprule
\multirow{2}{*}{\textbf{Model}}  &  \textbf{\%Novel}  &  \textbf{ Length }    \\

& \textbf{1/2/3 grams}&  \textbf{min/avg/max} \\ 
\midrule

Human-refer. & 37.00/83.18/98.16 & 23/46/99 \\ \midrule

BART-Large & 36.91/79.67/95.11 & 28/45/85 \\ 
+ specificity & 35.27/77.29/93.18 &  26/42/85 \\ \midrule

GPT-Human & 27.71/74.65/94.40 & 19/35/68 \\ 
+ specificity &29.10/74.19/94.02 &  15/38/66 \\ \midrule
GPT-one-shot & 31.43/77.59/95.39 & 15/35/66 \\
\bottomrule
\end{tabular}
\caption{Percent of novel n-grams and length statistics in  abstractive summaries.}\label{tab:abst_summ_analyze}
\end{table}

\subsection{Abstractive Summarization}
We report our abstractive summarization results in Table \ref{tab:abst_summ}.   Our results demonstrate that incorporating \textit{specificity markers} into the input achieved the best fine-tuned BART baseline performance by a small margin (rows 1 vs. 2; p < 0.05 for R-L). 
Unlike the results for the prior two extractive summarization tasks, none of the 
GPT-based LLMs could match the performance of 
more traditional methods, 
with respect to ROUGE and BERTScore. 
To examine whether characteristics of the LLM summary output might be a factor in their poor ROUGE performance,  Table \ref{tab:abst_summ_analyze} shows the percentage of novel n-grams present in each summary, as well as the average summary length.
The novelty figures indicate that LLMs generally have a lower proportion of novel n-grams compared to fine-tuned models, which maximize generating summaries that follow human-written summaries (high in novelty). However, incorporating one-shot learning improves n-gram novelty in LLMs, showing that providing even one example enhances the generation of novel outputs. Additionally, the abstractive summaries generated by LLMs are generally shorter than human-written summaries and those produced by fine-tuned models, potentially contributing to lower ROUGE.

Finally, Table \ref{tab:factual} shows the results of the 
factuality evaluation using  $\textsc{SummaC}$. For computing off-the-shelf entailment scores, we utilize the AlBERT model, which was fine-tuned on an entailment dataset as described in \citet{schuster2021get}. 
Our results indeed highlight the challenge and limitations of current factuality metrics when applied to this new type of data. 
The results indicate that \textit{GPT-Human} demonstrates the highest level of overall agreement with the input reflections, surpassing even the human-written summaries at both the sentence-level (\textit{GPT-Human +specificity}) and document-level resolutions. 
This finding is surprising since 
we guaranteed the quality of our human-written summaries by providing annotator training, as detailed in Section \ref{subsec:data_collection}. Therefore, a higher average SummaC score for the summaries generated by chatGPT does not necessarily indicate a more factual summary when compared to the human-written ones.\footnote{See Appendix \ref{subsec_app:summac_limit} for an illustrative example.}

We postulate that this disparity arises as our reflections may contain individual words or phrases rather than complete sentences, thereby deviating from the training data of the 
entailment model, which consists of complete sentences \cite{N18-1101}. 
Therefore, a thorough qualitative analysis is crucial to assess the factual accuracy of generated summaries. Novel factuality metrics tailored to reflective writing and similar summarization domains are promising avenues for future research.

\begin{table}[]
\small
\centering
\begin{tabular}{l|c|c}
 \toprule
\multirow{2}{*}{\textbf{Model}}  &  
\multicolumn{2}{c}{$\textbf{\textsc{SummaC}}$ $\uparrow$}

     \\

& \textbf{Sentence}&  \textbf{Document} \\ 
\midrule

Human-reference & 0.25 & 0.22 \\ \midrule

BART-Large & 0.25 & 0.21\\ 
+ specificity & 0.25 &  0.22 \\ \midrule

GPT-Human & 0.26 & \textbf{0.31} \\

+ specificity & \textbf{0.27} & 0.26 \\ \midrule

GPT-one-shot & 0.26 & 0.26 \\
\bottomrule
\end{tabular}
\caption{Factuality scores 
 based on $\textsc{SummaC} (\uparrow$:  higher means better entailment).}\label{tab:factual}
\end{table}

\section{Broader Impact of the Dataset}\label{sec:broader_impact}

This paper focused on introducing and utilizing the \textsc{ReflectSumm} dataset 
to develop and evaluate benchmark models for three summarization tasks.  For the NLP community, the dataset can enable the creation of new benchmarks for other tasks by harnessing the rich metadata to be released with the dataset.
For instance, researchers can use the student demographics to work on 
analyses of potential fairness or equity issues  \cite{Dash2019fairness} in summarization and build fairness-oriented summarization models. 
Initial explorations indicate that {\sc ReflectSumm} shows promise for this purpose, as it reveals a difference in the distribution of reflections along the gender dimension between the extractive summaries and the entire 
dataset.\footnote{There are more male-written reflections in extractive summaries (57\%) than in the overall distribution (52\%), verified with chi-square test.}

Moreover, new ways to use the manually annotated specificity annotations
beyond those presented here can
bridge summarization and NLP research on specificity prediction more broadly \cite{jessy-2015-specificity,Gao_Zhong_Preoţiuc-Pietro_Li_2019}. 
Also, our tasks and models could enable new downstream functionalities in educational technologies already collecting 
reflections \cite{Fan2015COURSEMIRROR,10.1145/3448139.3448166} such as generating recommended readings and explaining confusing concepts based on summary output. 
Learning scientists may also find our dataset valuable for monitoring the growth of students across a semester by analyzing reflections for the same course over time.   On average, students contributed course reflections to 42 percent of the lectures throughout the semester.  The longitudinal aspect of our data might also be used to define new summarization tasks.
\vspace{-10pt}
\section{Conclusion and Future Work}
We present \textsc{ReflectSumm}, a new dataset designed for course reflection summarization that includes multiple summarization tasks. The dataset provides specificity annotations of 
reflections and metadata on user demographics. To demonstrate its utility, we  
benchmarked the dataset. Our results demonstrated how the benefits of pretrained and finetuned language models,  large language models, reflection specificity, and one-shot learning techniques could vary significantly across different summarization tasks, shedding light on the nuanced advantages of these approaches.
\vspace{-10pt}
\paragraph{Future Work} We foresee numerous future directions that can be built upon the dataset's rich information coverage. 
For instance, we plan to enhance the efficacy of LLMs by investigating improved prompting techniques and developing more appropriate evaluation metrics that align with the 
nature of students' reflective writing. Our benchmark results also uncover that prompts modeled on human summarization guidelines are insufficient and that it remains challenging to incorporate best supervised examples and/or specificity into the summarization tasks. Additionally,  extending our dataset across different domains is necessary and beneficial for future research. We have collected data from two psychology courses (new domain) from a Canadian institute (outside the US) and a few Math and Mechanical Engineering courses from US institutes. We plan to add those as future additions to the \textsc{ReflectSumm} dataset. 
The number of extractive phrases or reflections should be dynamically adjusted for the extractive summarization tasks, as real-world scenarios vary across different lectures and subjects. Our analysis reveals 75 of the 782 lectures have 10 or fewer students. In these cases, the number of shared topics can be fewer than 5. One future direction would be a dynamic system that utilizes topic models to handle different-sized reflection collections.


\section*{Limitations}
For experiments involving the recent advanced LLMs, we created fairly straightforward benchmarks. Future work will need investigations of methods such as prompt optimization to better unleash the capability of ChatGPT and other LLMs \cite{qin2023chatgpt, bang2023multitask}. More supervised extractive/abstractive summarization models are yet to be added, and we encourage researchers to help contribute to the benchmarks. We also acknowledge that the released dataset may be of a narrow scope of subjects, including only engineering and science-related courses. With the released protocol, we envision extending the dataset to cover courses from the liberal arts and other backgrounds and facilitate the overall quality of teaching and education.  We also have not yet 
incorporated human assessments to gauge the quality
of various system outputs and explore the similarities
or differences between human-authored
and AI-generated outputs. In this resource paper,
our primary focus is to introduce the details
of the proposed dataset alongside showcasing its
tasks through benchmarking experiments carried
out on prevailing domain-specific models. Our experimental exploration, though not as focused on the intricacies of modeling as typically seen in modeling-based summarization
papers, offers a comprehensive
insight into the practical performance and applicability
of different models. It  establishes a
robust foundation for further investigations into  course reflection summarization.


\section*{Data Statement}
We are dedicated to making the dataset, including metadata such as demographic information and specificity scores, accessible to the public.  Furthermore, we pledge to provide supplementary materials instrumental in crafting this dataset and conducting experiments. These materials, such as the \textbf{annotation guidelines}, the \textbf{list of prompts} used for LLM experiments with GPT, and \textbf{examples of our evaluation metrics}, can be accessible at \url{https://github.com/EngSalem/ReflectSUMM} and in the Appendices.

\section*{Ethics Statement}
Safely using user demographic information becomes increasingly important in the current era. We carefully filtered the metadata that may leak the student's personal information, including emails, first and last names, and other attributes in the released version. We also only included data from students who explicitly provided consent.

\section*{Acknowledgement}
The research reported here was supported in part by the Institute of
Education Sciences, U.S. Department of Education, through Grant
R305A180477, and the National Science Foundation through Grants
2329273 and 2329274. The opinions expressed are those of the authors and do not represent the views of the U.S. Department of Education or the National Science Foundation. We want to thank the members of the Pitt PETAL group, Pitt NLP group, and anonymous reviewers for their valuable
comments in improving this work.
\section{References}\label{sec:reference}

\bibliographystyle{lrec-coling2024-natbib}


\appendix
\section{Human Annotation Details}

\subsection{Annotation Guidelines}\label{sec:appendix_annotation_guideline}
\begin{figure*}[h]
  \centering
  \includegraphics[width=\textwidth]{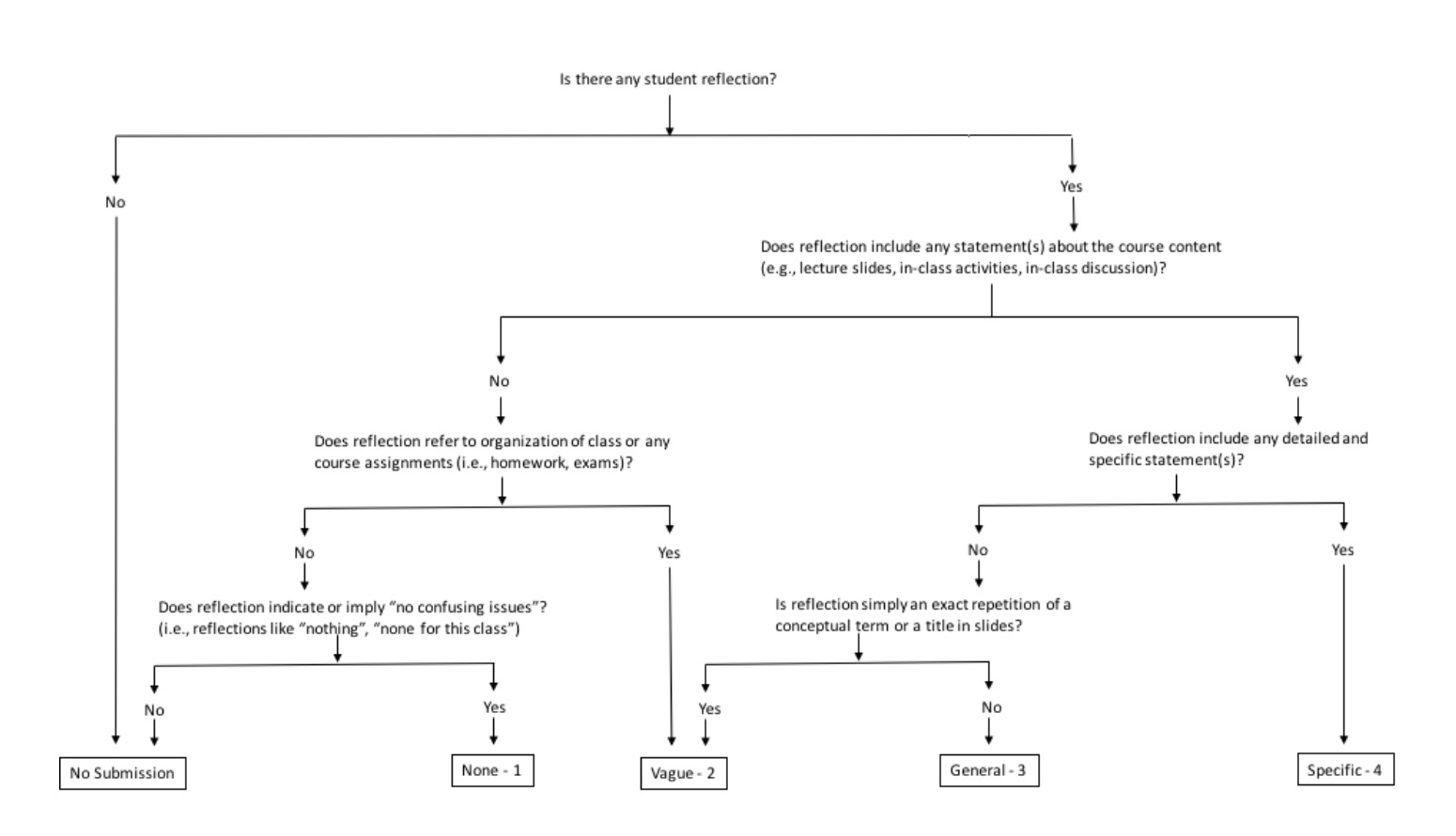}
  \caption{Human annotation guideline on the reflection specificity.}
  \label{fig:reflect_anno}
\end{figure*}

\begin{figure*}[ht!]
  \centering
  \includegraphics[width=\textwidth]{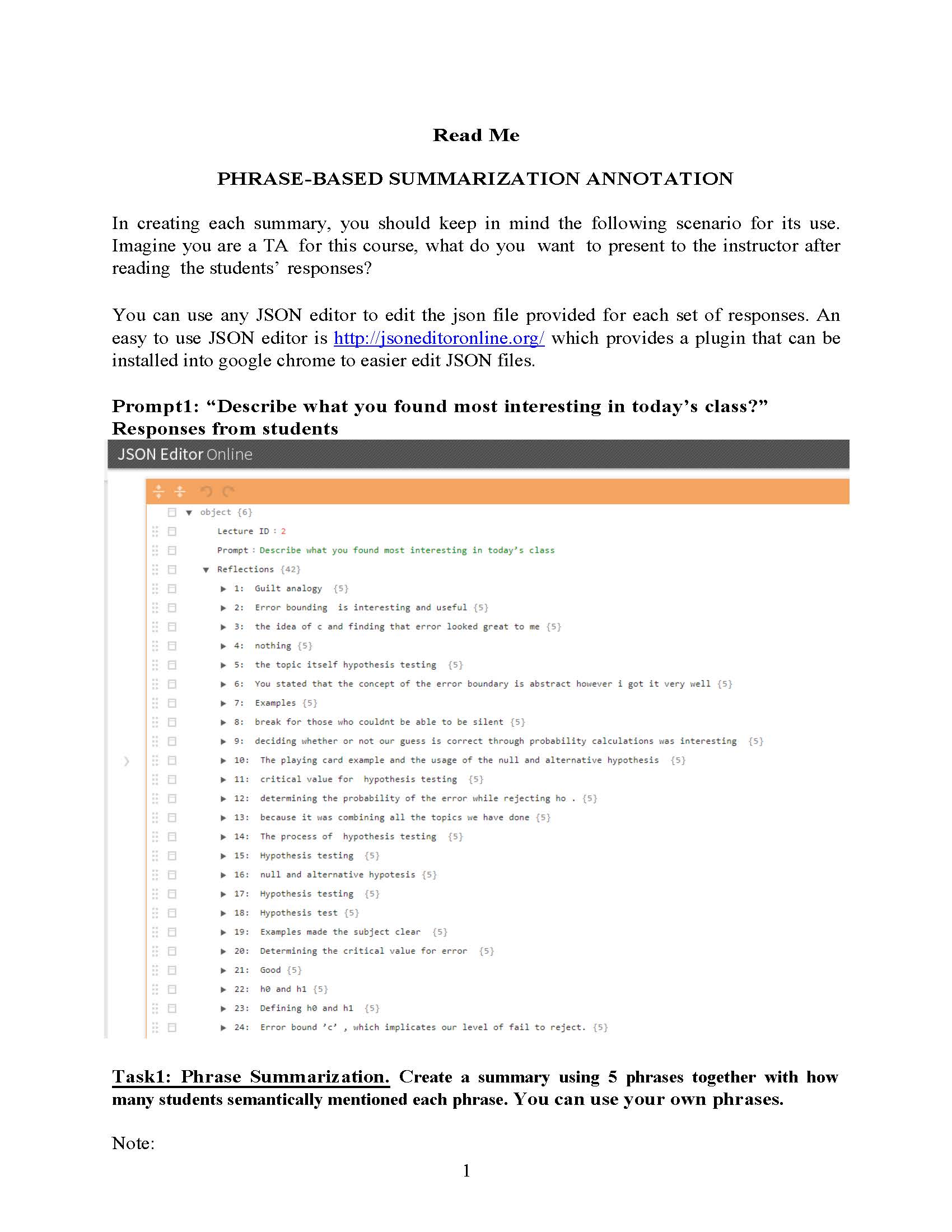}
  \caption{Human annotation guidelines  1/4.}
  \label{fig:annotation_1}
\end{figure*}

\begin{figure*}[t]
  \centering
  \includegraphics[width=\textwidth]{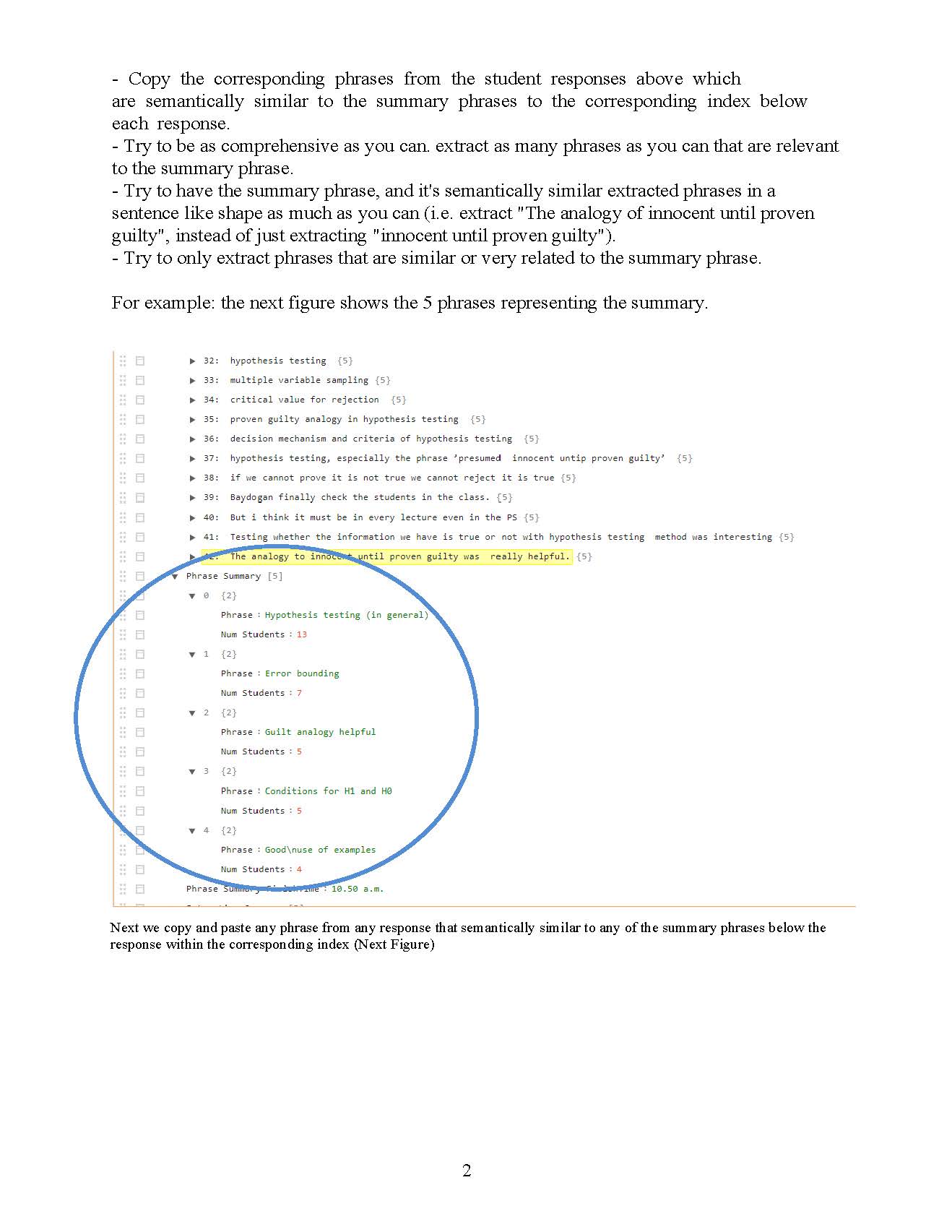}
  \caption{Human annotation guidelines  2/4.}
  \label{fig:annotation_2}
\end{figure*}

\begin{figure*}[t]
  \centering
  \includegraphics[width=\textwidth]{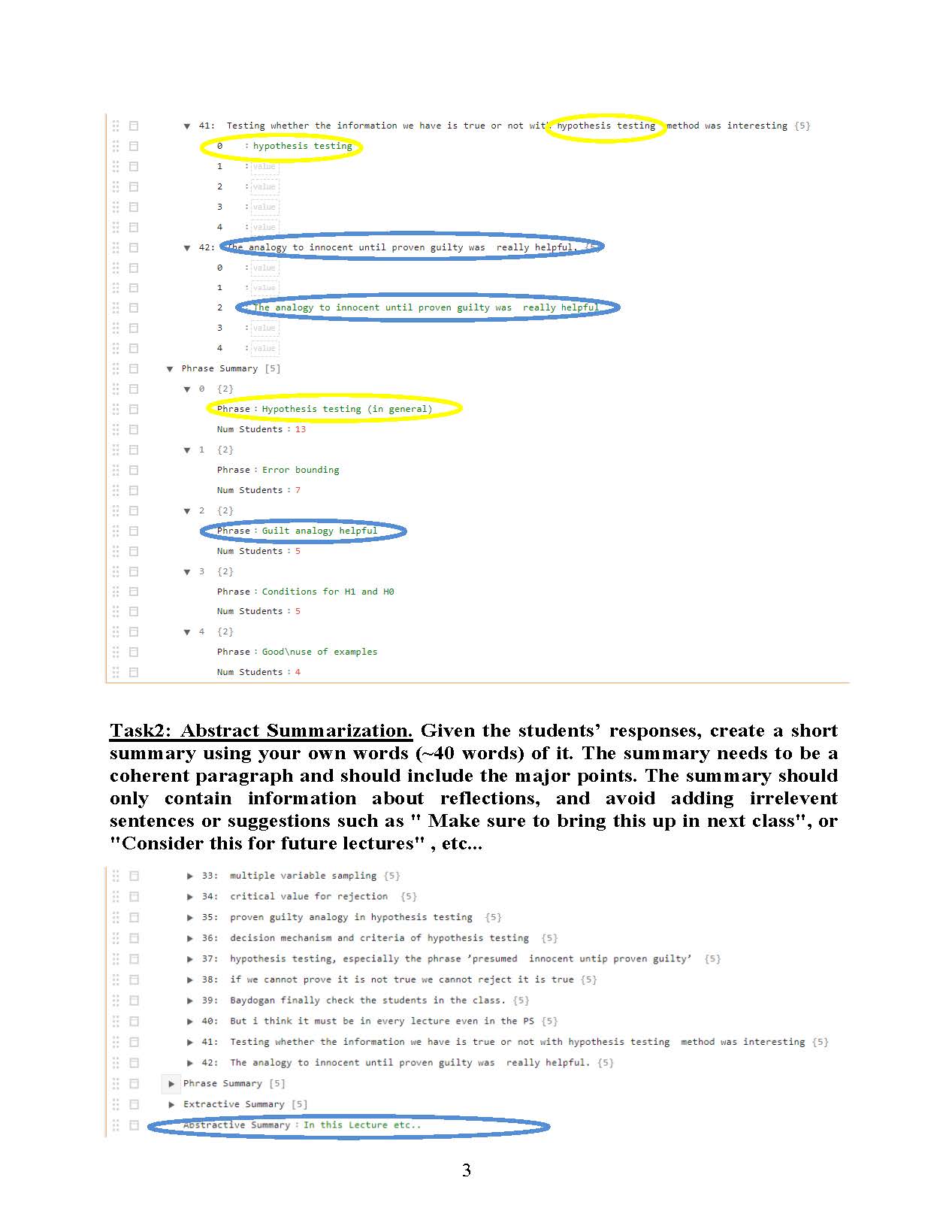}
  \caption{Human annotation guidelines  3/4.}
  \label{fig:annotation_3}
\end{figure*}

\begin{figure*}[t]
  \centering
  \includegraphics[width=\textwidth]{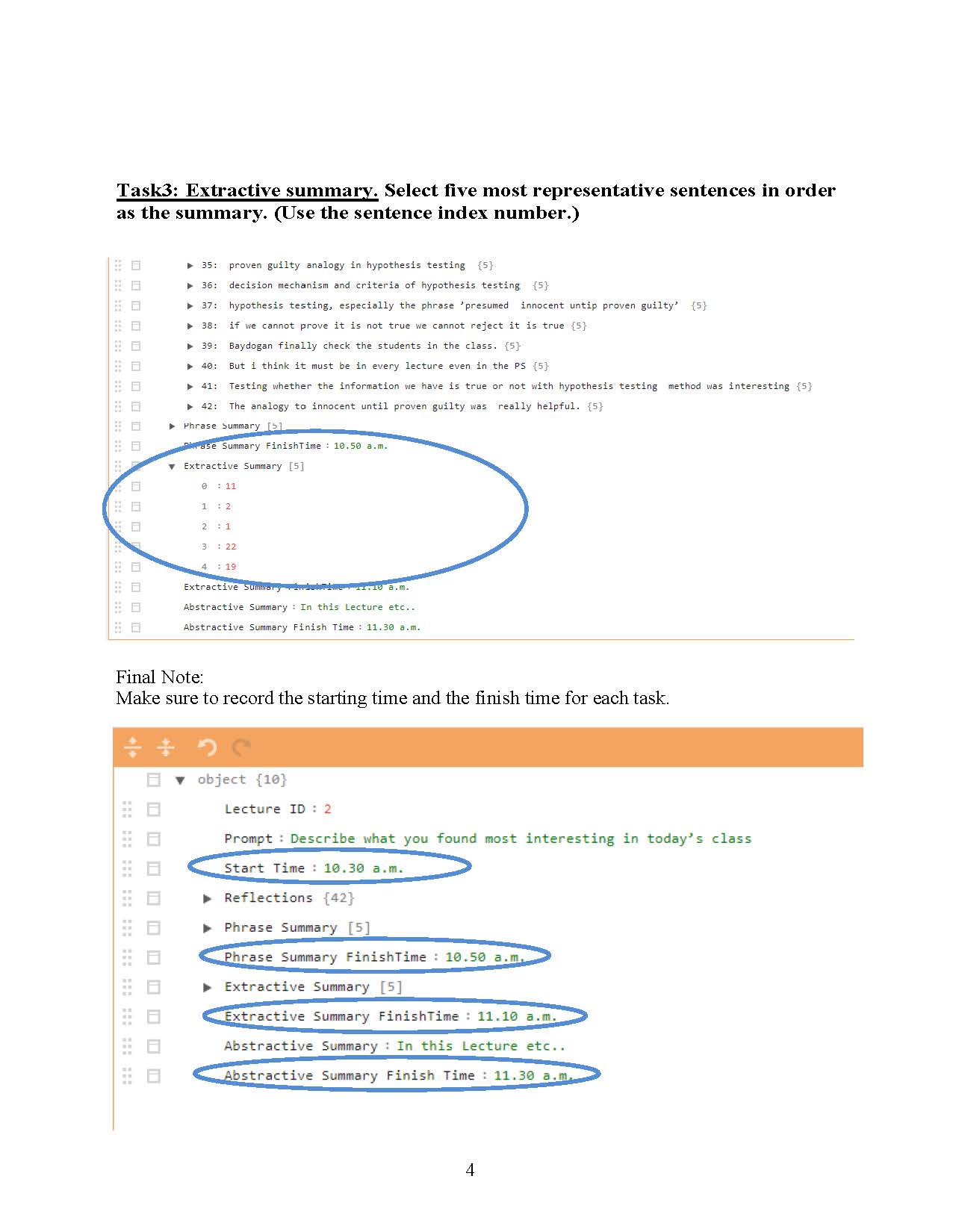}
  \caption{Human annotation guidelines  4/4.}
  \label{fig:annotation_4}
\end{figure*}

Figure \ref{fig:reflect_anno} shows the guideline we used for reflection specificity score annotation.  We provided the original human annotation guidelines for summarization tasks in Figures \ref{fig:annotation_1} to \ref{fig:annotation_4}.

\subsection{Rationales on Human Annotator Selection}
\label{sec:appendix_human_selection}
We worked with undergraduate students on this task for continuity and practicality. First, in prior motivating work from the learning sciences \cite{menekse2011reflection} and as noted in Section \ref{subsec:data_collection}, course TAs created summaries, so we wanted to keep a similar model, and we wanted to evaluate to what degree the NLP-generated summaries are similar to human-generated summaries. Second, practically, undergraduate students are more readily available and easier to recruit than course instructors. Importantly, our selection criteria for these individuals were not arbitrary; we chose students based on their academic majors. This ensured that they were familiar with the course content and had successfully met the course requirements themselves. 

It would be interesting to ask some experienced professionals to do the same task and compare the results between raters. This comparison could yield a richer understanding of summarization quality and style variances across different expertise levels.

\section{ChatGPT Prompts and BART's Markers}\label{sec:appendix_all_prompt}

\subsection{Extractive Prompts}\label{sec:appendix_sent_prompt_example}
We include the prompts for the extractive summarization task in Table \ref{tab:extractive_sent_prompt}, and the prompts for the extractive phrase summarization in Table \ref{tab:extractive_phrase_prompt}.

\begin{table*}[h]
\centering
\small
\begin{tabularx}{\textwidth}{c|X}
\toprule
Model & Prompt \\
\midrule
GPT-reflect & "\texttt{\textit{\small}reflections: \{\{reflections\}\} \newline Can you summarize the reflections, which are split by the special token $\|\_\|$, by extracting and selecting 5 original reflections from the split list?" } \\
GPT-reflect + specificity & "\texttt{\textit{\small}reflections: \{\{reflections\}\} \newline Can you summarize the reflections, which are split by the special token $\|\_\|$. Each reflection ends with a special marker -> and the specificity score in a range of 1-4, where 1 is the least specific and 4 is the most specific. Can you extract and select 5 original reflections from the split list by removing the ending "->" with the specificity score?"} \\
\bottomrule
\end{tabularx}
\caption{ChatGPT Prompts used for the extractive summarization task.}
\label{tab:extractive_sent_prompt}
\end{table*}

\begin{table*}[h]
\centering
\small
\begin{tabularx}{\textwidth}{c|X}
\toprule
Model & Prompt \\
\midrule
GPT-Human & "\texttt{\textit{\small}reflections: \{\{reflections\}\} \newline Create a summary using five phrases together with how many students semantically mentioned each phrase in parenthesis. You can use your own phrases.}" \\
GPT-noun phrase & "\texttt{\textit{\small}reflections: \{\{reflections\}\} \newline
Can you summarize the reflections by extracting and selecting five noun phrases?}"\\
GPT-Human + noun & "\texttt{\textit{\small}reflections: \{\{reflections\}\} \newline Create a summary using five \textbf{noun phrases} together with how many students semantically mentioned each phrase in parenthesis. You can use your own phrases.}" \\
\bottomrule
\end{tabularx}
\caption{ChatGPT Prompts used for the extractive phrase summarization task.}
\label{tab:extractive_phrase_prompt}
\end{table*}

\subsection{Abstractive BART + markers Example}
\label{sec:appendix_abst_summ}

During training, we use oracle markers, while we rely on model predictions during inference. We include an example in Table \ref{example_spec_Tok}.
\begin{table*}[h!]
\begin{tabular}{p{0.95\textwidth}}
\toprule
\textbf{Example}                                                                                                                                       \\ \hline
 \texttt{<high>} Energy conservation in particle physics with a gamma ray photon being split into an ... since I've never learned much about gamma rays in the past. \texttt{</high>}   ``\verb|\n|'' \texttt{<bad>} Nothing in particular today \texttt{</bad>}                                 \\ \bottomrule
\end{tabular}
\caption{An example of using markers for quality in our dataset, and all reflections are concatenated with new line symbols.\label{example_spec_Tok}}
\end{table*}

\begin{table*}[t!]
\centering
\small
\begin{tabularx}{\textwidth}{c|X}
\toprule
Model & Prompt \\
\midrule
GPT-Human & "\texttt{\textit{\small}reflections: \{\{reflections\}\}\} \newline Given the students’ responses, create a short summary with no more than $40$ words}"\\

GPT-Human + specificity & \texttt{\textit{\small}"reflections: \{\{reflections\}\}\} \newline Each reflection ends with a special marker -> and the specificity score in a range of 1-4, where 1 is the least specific and 4 is the most specific.   \newline Given the students’ responses, create a short summary with no more than $40$ words,  don't include the specificity scores in the summary."} \\
GPT-oneshot & "\texttt{\textit{\small}reflections: \{\{oneshot reflections\}\} \newline \textit{\small} summary: \{\{oneshot summary\}\} \textit{\small} reflections: \{\{test reflections\}\} \newline \textit{\small} summary:}" \\
\bottomrule
\end{tabularx}
\caption{ChatGPT Prompts used for the abstractive summarization task.}
\label{tab:abst_summ_prompts}
\end{table*}

\subsection{Abstractive Summarization Prompts}
\label{app:abst_prompts}

To ensure a fair comparison with models that utilized BART-large, which underwent fine-tuning on reference summaries generated by human annotators, where the annotators were instructed to create approximately $40$ word summaries, we followed a similar approach when providing instructions to GPT-Human. This was done to maintain consistency in the experimental setup and enable a meaningful comparison between the two models.
For the one-shot setting, we rely on in-context learning by providing an example from the reference summaries annotated by human annotators. Table \ref{tab:abst_summ_prompts} shows the prompts we used in each setting.  

\clearpage
\onecolumn
\section{Evaluation Metrics}
\subsection{Extractive F1 Metrics}
\label{appendix:partial_f1_illustration}
Besides the classical evaluation metrics including ROUGE and BERTScore, we additionally report the reflection-level exact match F1 and partial F1s. The intuition is to measure the selection accuracy of a given system provided with five human-annotated choices.  As shown in Table \ref{tab:partial_F1_example}, the GPT models can sometimes extract a portion of the original reflections, even with a carefully designed prompt. Using a too-strict exact matching would make the F1 score unfaithfully low, so we instead consider the partial matching on the sentence level and compute the partial match F1 accordingly in the main results.

\begin{table*}[h]
\centering
\begin{tabularx}{\textwidth}{X}
\toprule
\textit{Oracle Reflections} \\
\midrule
1. Three Sum example was very confusing, so it would be nice to go over it again. Also just more practice problems with determining the runtime of functions. And lastly just reiterating the formal definitions of the Big O family \\
2. The triple loops example and finding the runtime I got really lost in how we determine how to segment the loops into pieces and also how to calculate each individual output. Also why O(n) is the same as O(n+1) ie why the +1 is dropped from the calculation.  \\
3. I found it hard to follow some of the math done to analyze code examples, as well as towards the beginning with the formal definitions of the big o family.\\
4. More practice with determining run-times would be useful as well as more practice with the Big-O family functions\\
5. Mathematically modeling runtime was somewhat confusing as was determining the cost of executing each statement.\\
\midrule
\textit{GPT-reflect output} \\
\midrule
1. \underline{Three Sum example was very confusing, so it would be nice to go over it again.} \\
2. \underline{Just more practice problems with determining the runtime of functions.} \\
3. The triple loops example and finding the runtime I got really lost in how we determine how to segment the loops into pieces and also how to calculate each individual output.\\
4. I found it hard to follow some of the math done to analyze code examples, as well as towards the beginning with the formal definitions of the big o family.\\
5. Mathematically modeling runtime was somewhat confusing as was determining the cost of executing each statement.\\

\bottomrule
\end{tabularx}
\caption{An example, the first two GPT model outputs came from the same oracle reflection, and the third GPT model sentence belongs to the second oracle reflection. If we only consider the exact matching, the true positive count would be 2, while partial F1 considers the true positive as 4, which is more realistic and relaxed.}
\label{tab:partial_F1_example}
\end{table*}
\twocolumn
\subsection{Factuality Evaluation}
\label{app:factuality_evaluation}

In order to emphasize the difficulties associated with utilizing QA-based evaluation metrics for factuality assessment, we present an example in Table \ref{tab:entity_qa} that showcases the challenges encountered in entity extraction, which is a crucial component of the question generation module. Consequently, we opted not to incorporate QA-based factuality evaluation in our approach, acknowledging the limitations and complexities associated with this methodology.

\begin{table}[h!]
\begin{tabular}{p{0.45\textwidth}}
\toprule
\textbf{Input reflections}                 \\
\midrule
$\bullet$ Combinatorics. \\
$\bullet$ Understanding that the sum of two integers is still less then or equal to the product of twice the maximum of the set of integers being added was a little mind blowing since I was struggling understanding that for awhile.
Using the product rule to consider abstract examples like possible injective functions. \\
$\bullet$ Combinatorics seems really cool and it seems like it has a lot of real world and CS based applications.
the visual explanation of the binary trees was more helpful than my reading about them so that was interesting 
The combinatorics was a topic that I feel relatively comfortable with already. It seems straightforward and easy to comprehend as of now. \\
$\bullet$ The idea that induction has so many applications is interesting to me. \\ 
$\bullet$ Combinatorics. \\ 
\hline                            
\textbf{Extracted entities} 
\\ \midrule 
\textbf{Entity} two: \textbf{Type:} number
\\ \hline
\textbf{Extracted Nounphrases}
\\ \midrule
\textbf{Nounphrases:} Combinatorics,CS, sum, integers, two integers, sum of two integers , applications, trees, search trees ... \\
Shortened for space
\\ \bottomrule
\end{tabular}
\caption{Examples of the extracted entities and noisy nounphrases for a given input set of reflections.\label{tab:entity_qa}}
\end{table}

\subsection{Entailment Based Metrics Limitation}
\label{subsec_app:summac_limit}
To better understand the limitations of entailment methods in our domain-specific summarization task, Table \ref{tab:summac_example} presents an example of \textsc{SummaC} scores applied to a human-reference summary. The zero-shot score of $0.07$ indicates poor factuality. Additionally, the sentence-level \textsc{$SummaC_{conv}$} score of $0.21$ indicates poor factuality value (second lowest bin), which is counter-intuitive given that our human-reference summaries are accurate as shown in the example. This highlights the constraints of entailment-based metrics and emphasizes the need to explore factuality metrics tailored to the nature of students reflections.

\begin{table}[h!]
\begin{tabular}{p{0.45\textwidth}}
\toprule
\textbf{Input reflections}                 \\
\midrule
$\bullet$ One thing I found interesting was how many categories of machine learning there are. \\
$\bullet$ What discrete variables were and how they can be classified.\\
$\bullet$ None. \\
$\bullet$ The idea of discrete and continuous labels was most interesting. \\
$\bullet$ Supervised and unsupervised learning as well as discrete and continuous labels and how they all related to one another. \\
$\bullet$ The process for both categorization and classification. How one is based on context or perceived similarity and the other is a systematic arrangement of entities. \\
$\bullet$ How to categorize things as continuous or discrete. \\
$\bullet$ Different categories of machine learning. \\
$\bullet$ The algorithm systems for the ways the algorithms group different things on the way they identify the patterns. \\
$\bullet$ The relationship between unsupervised and supervised deep learning. \\
$\bullet$ I was intrigued by why discrete meant classification and how those 2 worked together was very interesting.
$\bullet$ Supervised vs unsupervised learning. \\
\midrule
\textbf{Human-reference summary}  \\
\midrule
Students enjoyed learning about the differences between supervised and unsupervised learning. Along with that, they also enjoyed learning about the different categories in Machine Learning and the different categorization and classification methods. \\
\midrule
\textsc{$SummaC_{conv}$} : $0.21$ \\
\midrule
\textsc{$SummaC_{zs}$}: $0.07$ \\
\bottomrule
\end{tabular}
\caption{\textsc{SummaC} scores on a human-reference summary example \label{tab:summac_example}}
\end{table}

\clearpage
\section{Experimental Details}
\subsection{PhraseSum Setups}\label{sec:appendix_phrasesum_detail}
We first reproduced the model from prior work \cite{luo-litman-2015-summarizing}. This model utilizes the spaCy toolkit \cite{Honnibal_spaCy_Industrial-strength_Natural_2020} to extract all noun phrases from student reflections, followed by the BERT-base model to extract the representation of these phrases. An unsupervised clustering method, KMedoid, is then employed to construct five clusters and extract their centroids as the final extractive phrases. Following \citet{luo-litman-2015-summarizing}, we named it \textbf{PhraseSum}. 

\subsection{Reflection Specificity Prediction}\label{sec:reflection_quality}
The goal of this task is to predict the specificity of student reflections. We first experimented with the prior model introduced by \citet{Magooda-etal-2022-improve}, which uses a DistillBERT model \cite{Sanh2019DistilBERTAD} followed by an SVM to predict scores on a 4-point scale. The baseline model was trained on the publicly available CourseMIRROR (CM) corpus\footnote{\url{https://engineering.purdue.edu/coursemirror/download/reflections-quality-data/.}}, consisting of 6,824 student reflections collected from four undergraduate classes (Chemistry (Chm), Statistics (ST), and Material Science (MSG1, MSG2)) at the end of each lecture. We retrained the model using a k-fold cross-validation setup. The QWK (Quadratic Weighted Kappa) scores improved from 0.624 to 0.689 when we trained the baseline model on our newly collected dataset. 

\subsection{Regular Expression for Phrase Cleaning}\label{appendix:regular_expression}
We have included the Python code snippets below, which demonstrate the regular expressions used in Section \ref{sec:result_and_analysis} to clean up the ``number of support'' for phrase outputs generated by GPT models. One example of applying the regular expressions together with some post-processing to remove the order numbers can be found in Table \ref{tab:re_example}. Since automatic metrics are sensitive to the n-gram wordings, removing the predicted values inside the parenthesis can make the comparison fair.

\begin{lstlisting}[language=Python,basicstyle=\footnotesize]

my_regex = re.compile(r"->(1|1.0|2.0
|2|3.0|3|4.0|4)|(\|\|)")
text = my_regex.sub('', text)
    
my_regex = re.compile(r"\(rated.*\)")
text = my_regex.sub("", text)
    
my_regex = re.compile(r"rated as a \d")
text = my_regex.sub("", text)
    
my_regex = re.compile(r"-> rating: (1|1.0|
2.0|2|3.0|3|4.0|4)")
text = my_regex.sub("", text)
    
my_regex = re.compile(r"- (1|1.0|2.0
|2|3.0|3|4.0|4)")
text = my_regex.sub("", text)
    
my_regex = re.compile(r"\(\d+ (student)
?s?\)")
text = my_regex.sub("", text)
    
my_regex = re.compile(r"\(\d+(student)
?s?\)")
text = my_regex.sub("", text)
    
my_regex = re.compile(r"\(\s+ (student)
?s?\)")
text = my_regex.sub("", text)
\end{lstlisting}

\begin{table}[ht!]
\begin{tabularx}{\columnwidth}{lX}
\toprule
\textbf{RegEx} & \textbf{Text}  \\ 
\midrule
Before & 1. Conditional probability examples (2) 2. Race with ties problem (1) 3. Challenge problems (1) 4. Sequences (1) 5. Disobeying conditional probability (1)\\
\midrule
After & Conditional probability examples \\
& Race with ties problem  \\ & Challenge problems\\  &  Sequences \\ & Disobeying conditional probability \\
 \bottomrule
\end{tabularx}
\caption{Examples of phrases processed using regular expressions in summaries generated by GPT models.\label{tab:re_example}}
\end{table}

\clearpage
\section{Output Examples}\label{sec:appendix_examples}

\paragraph{Extractive Summarization} Table \ref{tab:extract_sentence_example} and \ref{tab:extract_sentence_example_2} shows an example of different system outputs for the extractive summarization models. \textit{GPT-reflect} and \textit{GPT-reflect+specificty} extracted the full reflections with the modified prompt. The latter extracted more detailed reflections with the provided specificity hints. We also observe that MatchSum is more capable of extracting long reflections that align better with human preferences.

\paragraph{Extractive Phrase Summarization} Regarding the phrase summarization tasks, as shown in Table \ref{tab:extract_phrase_exmaple}, adding the noun prompt can make the generated phrases similar to the human written version. Yet, it is still unclear how the GPT models generate the student support count for each phrase (for both \textit{GPT-Human } and \textit{GPT-Human + noun}). 

\begin{table}[t!]
\centering
\footnotesize
\begin{tabular}{l}
\toprule
\textit{Oracle Phrase Summary} \\
\midrule
1. Ohm's Law (4) \\
2. Resistors/Resistance (6) \\
3. Circuits (4) \\
4. Power (2) \\
5. Real world applications (3) \\
\midrule
\textit{PhraseSum} \\
\midrule
1. the copper wire example \\
2. Ohm's Law \\
3. the resistors \\
4. real world incidences \\
5. how easy calculating resistance \\
\midrule
\textit{GPT-Human} \\
\midrule 
1. Ohm's Law and circuit analysis (8)\\ 2. Resistance in series and parallel (4) \\3. Battery voltage and electromotive force (3) \\4. Applying physics concepts to real life (2) \\5. Power and energy in circuits (2)\\

\midrule
\textit{GPT-noun phrase} \\
\midrule
1. Copper wire example \\2. Ohm's Law \\3. Circuits\\ 4. Batteries and resistance\\ 5. Practice problems \\
\midrule
\textit{GPT-Human + noun} \\
\midrule 
1. Circuit analysis (4) \\
2. Ohm's Law (3) \\ 3. Resistance calculation (2) \\ 4. Battery behavior (2) \\ 5. Power and energy (2) \\
\midrule
\textit{GPT-noun - one-shot} \\
\midrule 
1. Copper wire example \\
2. Ohm's Law \\
3. Circuits \\
4. Resistance calculations \\
5. Batteries and their characteristics \\

\bottomrule
\end{tabular}
\caption{An example of the extractive phrase summary and different model outputs.}
\label{tab:extract_phrase_exmaple}
\end{table}

\paragraph{Abstractive Summarization} In the context of abstractive summaries, Table \ref{tab:asbt_summ_example} provides examples of the generated summaries compared to those written by humans. Notably, models based on fine-tuned \textit{BART-Large} (BART-Large and Bart-Large + specificity) demonstrate a tendency to cover a broader range of topics compared to the reference summaries. For instance, the generated summary includes details about concepts such as "deep copy," "shallow copy," and specific implementation and usage of the "BinaryNode." Conversely, \textit{GPT-Human} tends to produce shorter summaries that do not delve into deeper details, aligning with our analysis indicating that \textit{GPT-Human} summaries are generally shorter compared to the fine-tuned BART models.  \textit{GPT-one-shot} generates summaries that encompass interesting topics related to confusion topics. We hypothesize that this might be due to the random selection of examples for one-shot learning, which could lead the model to capture intriguing aspects of the lecture content.

\begin{table*}[h]
\centering
\small
\begin{tabularx}{\textwidth}{X}
\toprule
\textit{Oracle Reflections} \\
\midrule
1. The idea of grounding was confusing for me. I feel like seeing some more examples would be helpful.\\
2. Problem 27.54 with the car motor problem was confusing. I did not understand what the starting motor does. Does it withdraw or add voltage? Does it act as a resistor or a battery? Otherwise, your lectures were clear and concise as always. \\
3. I am still a bit confused about why the voltage was negative in the first given example even though the voltage coming out of the battery is positive. \\
4. Current for the last top hat circuit question that was graded. \\
5. Re-drawing circuit diagrams with grounds seems to get complicated when multiple branches and resistors/capacitors are involved. \\
\midrule
\textit{LexRank} \\
\midrule 
1. Some of the ideas about grounding were confusing, especially with the top hat, but I think I have the general idea that a grounded circuit has a voltage drop of 0.\\
2. Some of the ideas about grounding were confusing, especially with the top hat, but I think I have the general idea that a grounded circuit has a voltage drop of 0. \\
3. The most confusing thing for me was when we went over the 2 different cases of grounding. I just found it confusing when going over the formulas like whether to add or subtract $IR$.\\
4. I found grounding to be very confusing; what does it mean for a circuit to be grounded? \\
5. Conceptually, I was a little confused how you can have a ground before a resistor, making the voltage zero, and that doesn’t impact the energy continuing on. Mathematically I can see why this is true, but conceptually it’s a bit confusing.\\
\midrule
\textit{GPT-reflect} \\
\midrule
1. The idea of grounding was confusing for me. I feel like seeing some more examples would be helpful.\\ 
2. I'm confused about how the spot where we place a ground affects how we measure voltage.\\ 
3. The most confusing thing for me was when we went over the 2 different cases of grounding.\\ 
4. The only thing that confuses me slightly is the sign associated with $IR$ when going in the flow of charge vs against the flow of charge. Also, how would we determine where the charge flows?\\ 
5. Finding the emf was confusing. \\
\midrule
\textit{GPT-reflect + specificity} \\
\midrule 
1. I am still a bit confused about why the voltage was negative in the first given example even though the voltage coming out of the battery is positive.\\ 
2. Some of the ideas about grounding were confusing, especially with the top hat, but I think I have the general idea that a grounded circuit has a voltage drop of 0.\\ 
3. The conceptual questions during class were the hardest for me to understand because I kept getting the wrong.\\ 
4. It was really hard to understand the currents and why they were going in the direction they were.\\ 
5. The idea of grounding was confusing for me. I feel like seeing some more examples would be helpful.\\
\midrule
\textit{GPT-reflect one-shot} \\
\midrule 
1. The conceptual questions during class were the hardest for me to understand because I kept getting the wrong answers. \\ 
2. The idea of grounding was confusing for me. I feel like seeing some more examples would be helpful. \\ 
3. More examples of simplifying circuits would be helpful and a bit more explanation about grounding. \\ 
4. The most confusing thing for me was when we went over the 2 different cases of grounding. I just found it confusing when going over the formulas like whether to add or subtract $IR$. \\ 
5. The sapling question about finding emf and internal resistance has me stumped. If we could go over more of these that would be awesome.\\
\bottomrule
\end{tabularx}
\caption{An example of the difference between the oracle extractive summary and the generated model outputs (1/2).}
\label{tab:extract_sentence_example}
\end{table*}

\begin{table*}[h]
\centering
\small
\begin{tabularx}{\textwidth}{X}
\toprule
\textit{Oracle Reflections} \\
\midrule
1. The idea of grounding was confusing for me. I feel like seeing some more examples would be helpful.\\
2. Problem 27.54 with the car motor problem was confusing. I did not understand what the starting motor does. Does it withdraw or add voltage? Does it act as a resistor or a battery? Otherwise, your lectures were clear and concise as always. \\
3. I am still a bit confused about why the voltage was negative in the first given example even though the voltage coming out of the battery is positive. \\
4. Current for the last top hat circuit question that was graded. \\
5. Re-drawing circuit diagrams with grounds seems to get complicated when multiple branches and resistors/capacitors are involved. \\
\midrule
\textit{BERTSUM-EXT (cnndm)} \\
\midrule 
1. I am still a bit confused about why the voltage was negative in the first given example even though the voltage coming out of the battery is positive. \\
2. Problem 27.54 with the car motor problem was confusing. I did not understand what the starting motor does. Does it withdraw or add voltage? Does it act as a resistor or a battery? Otherwise, your lectures were clear and concise as always.\\
3. The idea of grounding was confusing for me. I feel like seeing some more examples would be helpful.\\ 
4. Some of the ideas about grounding were confusing, especially with the top hat, but I think I have the general idea that a grounded circuit has a voltage drop of 0. \\
5. I found grounding to be very confusing; what does it mean for a circuit to be grounded? \\
\midrule
\textit{BERTSUM-EXT (ft. \textsc{FewSum Amazon})} \\
\midrule 
1. I found grounding to be very confusing; what does it mean for a circuit to be grounded? \\
2. Why it is useful to have a reference point. I am confused about what exactly delta V is. \\
3. The most confusing thing for me was when we went over the 2 different cases of grounding. I just found it confusing when going over the formulas like whether to add or subtract $IR$. \\
4. I'm confused about how the spot where we place a ground affects how we measure voltage. \\
5. The sapling question about finding emf and internal resistance has me stumped. If we could go over more of these that would be awesome. \\
\midrule
\textit{BERTSUM-EXT (ft. \textsc{ReflectSumm})} \\
\midrule 
1. Conceptually, I was a little confused how you can have a ground before a resistor, making the voltage zero, and that doesn't impact the energy continuing on. Mathematically I can see why this is true, but conceptually it's a bit confusing. \\
2. I am still a bit confused about why the voltage was negative in the first given example even though the voltage coming out of the battery is positive. \\
3. Finding the emf was confusing. \\
4. Why it is useful to have a reference point. I am confused about what exactly delta V is. \\
5. Figuring out how grounding will affect the potential. \\
\midrule
\textit{BERTSUM-EXT (ft. \textsc{ReflectSumm}) + Specificity} \\
\midrule
1. I am still a bit confused about why the voltage was negative in the first given example even though the voltage coming out of the battery is positive.  \\
2. It was really hard to understand the currents and why they were going in the direction they were. \\
3. Current for the last top hat circuit question that was graded. \\
4. Some of the ideas about grounding were confusing, especially with the top hat, but I think I have the general idea that a grounded circuit has a voltage drop of 0. \\
5. Figuring out how grounding will affect the potential. \\
\midrule
\textit{MatchSum} \\
\midrule 
1. Some of the ideas about grounding were confusing, especially with the top hat, but I think I have the general idea that a grounded circuit has a voltage drop of 0. \\
2. The clicker questions about grounding. \\
3. Problem 27.54 with the car motor problem was confusing. I did not understand what the starting motor does. Does it withdraw or add voltage? Does it act as a resistor or a battery? Otherwise, your lectures were clear and concise as always. \\
4. Finding the emf was confusing. \\
5. I found grounding to be very confusing; what does it mean for a circuit to be grounded? \\
\bottomrule
\end{tabularx}
\caption{An example of the difference between the oracle extractive summary and the generated model outputs (2/2).}
\label{tab:extract_sentence_example_2}
\end{table*}

\begin{table*}[ht!]
\centering
\begin{tabularx}{\textwidth}{X}
\toprule
\textit{Human-reference} \\
\midrule
In this lecture, students were most confused with the BinaryNode class. They also require further instruction on methods to construct and implement a binary tree. Also, students were confused on the different traversal methods. \\
\midrule
\textit{BART-Large} \\
\midrule 
In this lecture, students were confused by the BinaryNode class and BinaryTree implementation. Students were also confused by BinaryNode's implementation and how it used BinaryNode objects to execute all its functions. Students also found it confusing to understand the difference between deep copy and shallow copy. They also were confused about the code for the different traversal methods. Lastly, they were confused on the binary tree diagram. \\
\midrule
\textit{BART-Large + specificity} \\
\midrule
In this lecture, students were confused by the BinaryNode and BinaryTree methods. Students also found the code for the different traversal methods confusing in how it used BinaryNode objects to execute all its functions. They also found it confusing to learn about the difference between deep copy and shallow copy. Students were also confused by the tree diagram and the last example shown in class. \\
\midrule
\textit{GPT-Human} \\
\midrule
The class covered topics such as deep copy vs shallow copy, Binary Tree implementation, BinaryNode objects, and tree traversal methods. Some students found these topics confusing, while others did not have any questions. \\
\midrule
\textit{GPT-Human + specificity} \\
\midrule 
Students found the implementation of BinaryNode and BinaryTree methods confusing, especially the difference between deep copy and shallow copy. Some were lost during the last example and had trouble understanding the constructor and copying a binary tree. \\
\midrule
\textit{GPT-oneshot} \\
\midrule
Students found binary tree implementation confusing, particularly the deep copy vs shallow copy in BinaryNode class and the BinaryTree build tree method. Some were unsure of the last example discussed. Tree traversal and terminology were interesting. \\
\bottomrule
\end{tabularx}
\caption{An example of the reference summary and different abstractive model outputs.}
\label{tab:asbt_summ_example}
\end{table*}

\twocolumn

\clearpage

\end{document}